\documentclass[lettersize,journal]{IEEEtran}
\usepackage{amsmath,amsfonts}
\usepackage{algorithmic}
\usepackage{algorithm}
\usepackage{array}
\usepackage[caption=false,font=normalsize,labelfont=sf,textfont=sf, justification=centering]{subfig}
\usepackage{textcomp}
\usepackage{stfloats}
\usepackage{url}
\usepackage{verbatim}
\usepackage{graphicx}
\usepackage{bm}
\usepackage{cite}
\usepackage{color}
\usepackage{CJKutf8}
\begin{document}

\title{ A prior information informed learning architecture for flying trajectory prediction }

\author{Xianda Huang, Zidong Han, Ruibo Jin, Zhenyu Wang, Wenyu Li, Xiaoyang Li, and Yi Gong
\thanks{Xianda Huang, Zidong Han, Zhenyu Wang, Xiaoyang Li, and Yi Gong are with the Southern University of Science and Technology, Shenzhen, China. Ruibo Jin and Wenyu Li are with the Chinese University of Hong Kong-Shenzhen, Shenzhen, China. The first two authors contributed equally to this work. Corresponding author: Yi Gong (gongy@sustech.edu.cn). }
\thanks{ }
}


\maketitle

\begin{abstract}
Trajectory prediction for flying objects is critical in domains ranging from sports analytics to aerospace. However, traditional methods struggle with complex physical modeling, computational inefficiencies, and high hardware demands, often neglecting critical trajectory events like landing points. This paper introduces a novel, hardware-efficient trajectory prediction framework that integrates environmental priors with a Dual-Transformer-Cascaded (DTC) architecture. We demonstrate this approach by predicting the landing points of tennis balls in real-world outdoor courts. Using a single industrial camera and YOLO-based detection, we extract high-speed flight coordinates. These coordinates, fused with structural environmental priors (e.g., court boundaries), form a comprehensive dataset fed into our proposed DTC model. A first-level Transformer classifies the trajectory, while a second-level Transformer synthesizes these features to precisely predict the landing point. Extensive ablation and comparative experiments demonstrate that integrating environmental priors within the DTC architecture significantly outperforms existing trajectory prediction frameworks.
\end{abstract}


\section{Introduction}
\IEEEPARstart{T}{he} proliferation of intelligent systems across aerospace and sports analytics—specifically those driven by advanced computer vision—has created a critical need for the accurate trajectory prediction of aerial targets, drawing substantial interest from both academia and industry \cite{ref1, ref2, ref3, ref4, ref5, ref6, ref7}. For example, notable applications encompass air traffic management within the aviation sector and motion tracking utilized in officiating during competitive sports events. Formally, trajectory prediction aims to estimate a target's dynamic state by modeling its spatiotemporal dynamics to forecast future behavior accurately. However, predicting the trajectories of flying objects remains a formidable challenge. Their states are governed by high-order, nonlinear dynamics that are highly sensitive to complex environmental variations, making accurate physical modeling exceedingly difficult. Furthermore, achieving high-precision predictions relies on extensive, high-fidelity trajectory datasets, which are both time-consuming and costly to acquire.

Current trajectory prediction methods generally fall into two paradigms: model-based and data-driven. Model-based approaches leverage kinematic models and boundary conditions to project future states \cite{ref8, ref9, ref10}. While they provide structured frameworks that are highly effective for short-term prediction, their computational complexity escalates dramatically with system dimensionality, severely limiting their scalability in complex scenarios. Conversely, data-driven approaches—particularly deep learning \cite{ref11, ref12, ref13}—excel at extracting nonlinear flying patterns directly from historical datasets. However, existing methods often neglect critical environmental priors and physical constraints, such as obstacle-impacted landing points. Furthermore, they demand massive volumes of high-quality, multi-camera training data, resulting in prohibitive collection and preprocessing costs.

In this paper, we propose a novel trajectory prediction method that integrates flight data with environmental priors to accurately forecast the landing points of tennis balls in real-world outdoor courts. Our pipeline begins with a custom data acquisition system featuring a single high-speed 2D camera (150–250 fps) and a professional ball launch machine. We employ YOLOv10 for precise ball detection, alongside edge and Hough line detection to extract critical court boundaries (e.g., corners and sidelines) as environmental priors. To process this integrated dataset, we introduce a Prior Information-Informed Dual-Transformer-Cascaded (PIDTC) architecture. Within this model, a first-level Transformer classifies trajectories using the environmental priors, while a second-level Transformer synthesizes these features to pinpoint the final landing coordinates. Extensive experiments validate the superior accuracy and effectiveness of our proposed approach.

The main contributions of our work are summarized as follows.
\begin{itemize}
\item{We propose a novel transformer-based model for flying object trajectory prediction. This architecture specifically targets the accurate forecasting of critical trajectory moments (e.g., landing points), addressing a major gap in existing data-driven approaches.}
\item{We construct a comprehensive trajectory dataset using a cost-effective, 2D monocular industrial camera setup to capture high-speed grayscale images. This methodology significantly reduces the hardware complexity and financial costs associated with conventional multi-camera acquisition systems.}
\item{We integrate environmental priors (e.g., court corners) with standard trajectory data to enrich the physical characterization of 2D flight paths. Extensive experiments validate that leveraging these enhanced features within our PIDTC architecture substantially outperforms existing baseline methods.}
\end{itemize}

The rest of the paper is organized as follows. Section II introduces the related work. Section III introduces the construction of the trajectory dataset. Section IV presents the proposed trajectory prediction model, outlining its theoretical foundations. Section V describes the experimental setup, followed by a thorough analysis of the experimental results obtained. Finally, Section VI concludes this paper.

\section{Related Work}
\subsection{Kinematic Model-based Trajectory Prediction}
Model-based trajectory prediction methods utilize kinematic models to project future states based on governing motion laws. Existing literature, predominantly focused on table tennis, typically involves two critical steps: establishing the kinematic model and determining boundary conditions.

The establishment of table tennis kinematic model primarily relies on high-order polynomial fitting \cite{ref14}, \cite{ref15}. The pioneering kinematic model for table tennis that incorporated spin effects was proposed in \cite{ref14} via a quintic polynomial, successfully enabling the prediction of table tennis trajectories. To enhance the accuracy of trajectory predictions, \cite{ref16} determined key parameters, such as the resistance coefficient and Magnus force coefficient, through visual measurements, subsequently integrating these coefficients into the kinematic model. Given the inherent complexity of the ball's motion, a single model often proves insufficient. Consequently, some researchers employ multiple models to capture the motion from various perspectives. In \cite{ref17}, two distinct table tennis kinematic models were proposed: the discrete model and the continuous model. The discrete model is employed for the state estimation of the flying ball, while the continuous model is tasked with predicting trajectories based on the ball's current state. A collision model was utilized in \cite{ref18} to estimate the ball motion parameters after the ball's impact, and subsequently integrated a kinematic model with motion parameters to predict post-collision trajectory.

Beyond the kinematic model, the determination of boundary conditions plays a crucial role in enhancing predictive performance. The boundary conditions primarily involves the initial states of the flight ball, such as position and velocity. Fourier series method was employed in \cite{ref19} and \cite{ref20} to fit the velocity variation values of the flight ball, then extract the initial states of the table tennis. Kalman filtering-based method was introduced to mitigate environmental noise. \cite{ref21} employed an extended Kalman filter to measure the spin state of table tennis and performed force analysis to establish its kinematic model. \cite{ref22} proposed trajectory prediction methods utilizing the unscented Kalman filter, which addresses the low estimation efficiency of the extended Kalman filter. 

Although the model-based trajectory prediction methods has demonstrated the ability to deliver high-precision outcomes in short-term forecasting tasks, there are obvious challenges associated with these methods: 1) The kinematic models present challenges in accurately predicting the influence of random factors on long-term forecasting tasks; 2) To accurately represent special trajectories, particularly collision points, it is essential to re-establish the collision model \cite{ref23}. This necessity inherently raises the modeling cost associated with these scenarios.

\subsection{Data-driven Trajectory Prediction}
\begin{table}[h]
\caption{Summary of data-driven trajectory prediction\label{tab:table1}}
\centering
\begin{tabular}{|c||c|}
\hline
The type of trajectory & Prediction model\\
\hline
Flying ball trajectory & RNN \cite{ref13}; LSTM \cite{ref11, ref12}, \cite{ref32}\\
\hline
Vehicle trajectory & Markov \cite{ref24}; LSTM \cite{ref28};\\& Transformer \cite{ref35}\\
\hline
Ship trajectory & GRU\cite{ref29}, \cite{ref38}; LSTM \cite{ref30}\\
\hline
\end{tabular}
\end{table}

\begin{figure*}[!t]
\centering
\subfloat[]{\includegraphics[width=3in]{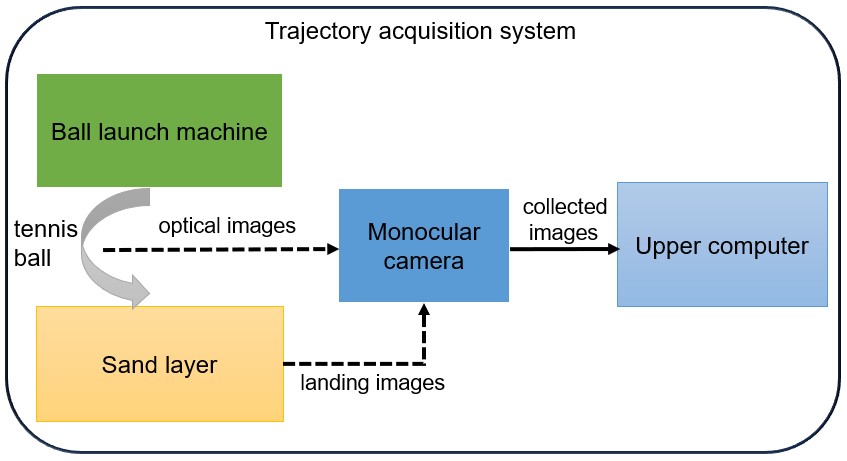}%
\label{fig1sub1}}
\hfil
\subfloat[]{\includegraphics[width=4in]{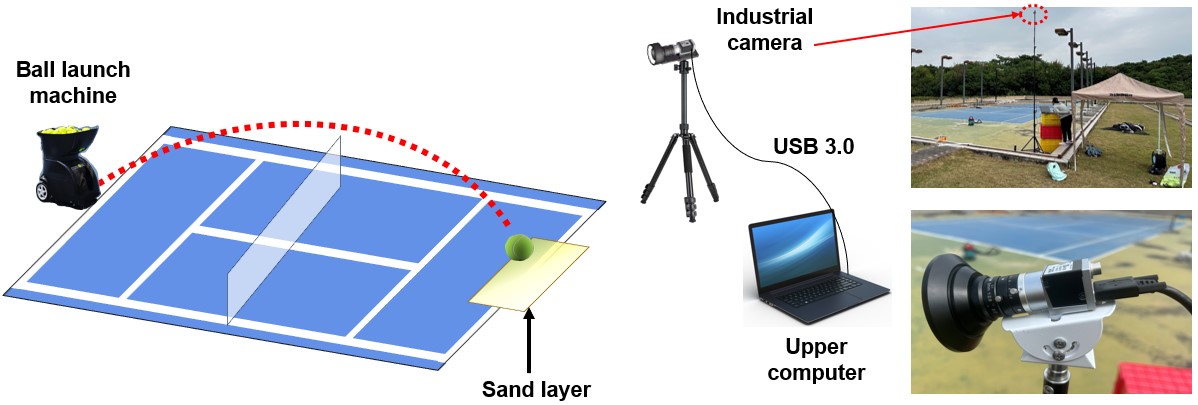}%
\label{fig1sub2}}
\caption{(a) The schematic overview of the trajectory data acquisition system. (b) Schematic diagram of the experimental scene.}
\label{fig1}
\end{figure*}

Data-driven trajectory prediction methods extract underlying kinematic patterns from historical datasets. While early statistical approaches like Hidden Markov Models \cite{ref24} and Bayesian inference \cite{ref25} suffered from limited accuracy, the advent of deep learning has significantly advanced the field.

Initially, neural networks were leveraged to enhance model-based trajectory prediction, including the estimation of a ball's spin state during flight \cite{ref26}, \cite{ref27}. Later, some researchers began to apply \textit{recurrent neural networks} (RNNs) for direct trajectory prediction \cite{ref28, ref29, ref30}. RNNs represent a specific category of neural architectures designed for processing sequential data, including foundational RNNs, \textit{gated recurrent units} (GRUs) and \textit{long short-term memory} networks (LSTMs) \cite{ref31}. For example, a structure of numerous LSTM units was introduced to fit the motion patterns, thus enabling long-term trajectory predictions for a flying ball \cite{ref11}. To further refine target detection capabilities within recognition systems, a cross-stage partial network was introduced, which enhanced the predictive performance of LSTMs \cite{ref32}. To predict the landing point of table tennis, an LSTM-based trajectory prediction model incorporating a mixture density network was proposed \cite{ref12}. A method utilizing two RNNs was suggested to forecast two trajectories that diverge at the collision point, thereby reducing the impact of the ball's collision \cite{ref13}. However, the iterative nature inherent to RNNs results in cumulative errors during long-term predictions.

To address the limitation of RNN, some other models were proposed for trajectory prediction. A \textit{convolutional neural network} (CNN)-based method was proposed to incorporate human pose information, thereby aiding the subsequent LSTM network in predicting table tennis landing point \cite{ref33}. A deep conditional generative model was proposed for trajectory prediction \cite{ref34}, which employs an encoder-decoder architecture to effectively mitigate the error accumulation problem commonly encountered with RNNs. Furthermore, the Transformer model, characterized by its analogous encoder-decoder architecture, has been successfully utilized in predicting some other types of trajectories \cite{ref35, ref36, ref37}. 

Incorporating reasonable prior information alongside suitable neural networks has emerged as a promising strategy for enhancing predictive performance \cite{ref38}. The integration of prior knowledge has been demonstrated to improve both the accuracy and training efficiency of neural networks across various domains, including pose estimation \cite{ref39}, image restoration \cite{ref40} and 3D reconstruction \cite{ref41}. In addition, for vehicle trajectory prediction, the lane-changing intention was used as prior information \cite{ref42}. For vessel trajectory prediction, the automatic identification system data was coded as the input \cite{ref43}, achieving a lower prediction error compared to the standard LSTM. The predicted course was integrated as a prior information in \cite{ref44} for the short-term prediction of vessel trajectory. 

Although data-driven methods demonstrate significant potential for predicting flying trajectories, there remains obvious challenges associated with these existing approaches: 1) They require large volumes of high-quality data for training, resulting in considerable collection cost; 2) Most of them only rely on the input trajectories, ignore other useful information, such as the prior environmental/contextual information; 3) Most of them neglect the critical points of the trajectory impacted by physical obstacles, such as landing points.

\section{Flying Trajectory Dataset Construction}

\begin{figure*}[!t]
\centering
\includegraphics[width=6in]{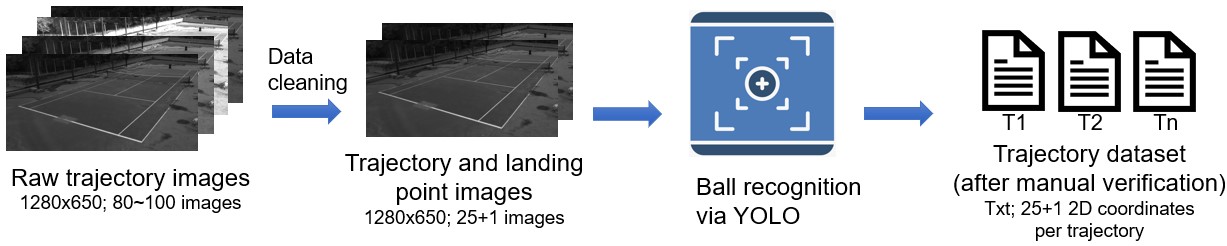}
\caption{Flowchart of dataset construction.}
\label{fig2}
\end{figure*}
\subsection{Data Acquisition System}
A schematic overview of the data acquisition system is illustrated in Fig. 1. A Jbotsports JW-05 ball launch machine is positioned at the center of the baseline, opposite the camera. To ensure comprehensive trajectory coverage, a Basler acA1920-155um industrial camera (equipped with a 5 mm wide-angle lens) is mounted on a 5-meter tripod at the court's corner. The camera captures images at 164 fps with a resolution of 1280×650 pixels. To minimize environmental interference, data collection was conducted exclusively under clear, calm weather conditions. The acquisition protocol proceeds as follows:
\begin{enumerate}
    \item System Calibration: The camera's field of view and exposure settings are optimized via a upper computer to ensure clear identification of both the court lines and the flying ball.
    \item Data Acquisition: The camera and ball machine are activated, continuously recording the flight path until the ball impacts the target sand layer.
    \item Scene Reset: After each valid recording, the sand layer is smoothed to erase the landing mark, preventing data contamination in subsequent trials.
\end{enumerate}

Due to the inherent mechanical variance of the launch machine, trajectories landing squarely within the designated sand area accounted for less than 20\% of the trials. Consequently, we curated a final dataset of 350 highly qualified trajectories from an initial pool of more than 2,000 recordings.

\subsection{Dataset Construction Method}
As illustrated in Fig. 2, the raw video data undergoes a systematic preprocessing pipeline—encompassing data cleaning, YOLOv10-based ball detection \cite{ref45}, and coordinate extraction—to finalize a dataset of 350 valid trajectories. The procedure is structured as follows:
\begin{enumerate}
\item{YOLOv10 Training: To ensure precise detection, the YOLO model was trained on 5,000 annotated images (split 4:1 for training and validation) over 300 epochs with a batch size of 16. Under our experimental lighting conditions, the model achieved a recognition accuracy exceeding 98\%.}
\item{Trajectory Data Cleaning: Capturing the precise moment of landing is challenging; therefore, the ball's initial bounce serves as the ground-truth landing indicator. For each sample, we extract the 25 flight frames immediately preceding this bounce, resulting in 25 trajectory points and 1 landing point per sequence.}
\item{Coordinate Extraction \& Verification: The trained YOLOv10 model detects the ball across all frames, saving the 2D spatial coordinates to text files. Finally, all outputs undergo manual verification to eliminate any anomalous detections.}
\end{enumerate}

\section{Trajectory Prediction Model}

\begin{figure*}[!t]
\centering
\includegraphics[width=6in]{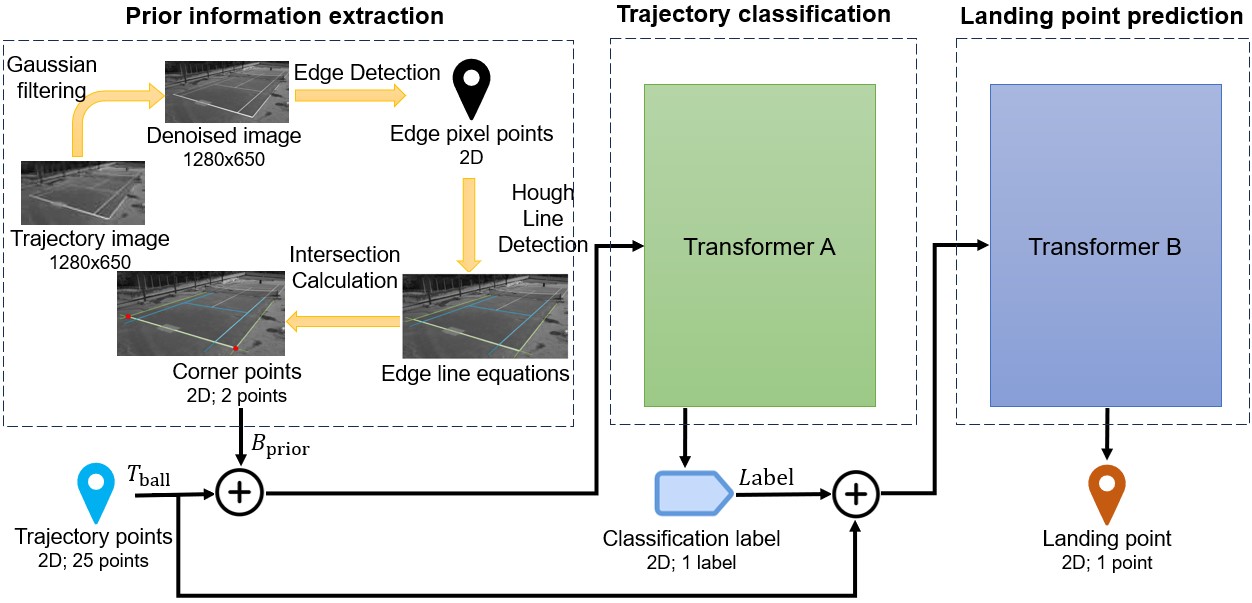}
\caption{ Overview of the trajectory prediction model - PIDTC.}
\label{fig3}
\end{figure*}

This section details the proposed prediction model, illustrated in Fig. 3. The architecture comprises two core components: a prior information extraction module and a dual-Transformer-cascaded prediction network. Within the DTC framework, the first-level Transformer acts as a trajectory classifier—determining whether the landing point will fall ``in'' or ``out'' of the court boundaries—while the second-level Transformer leverages this classification to precisely forecast the final landing coordinates.

\subsection{Prior Information Extraction Module}

As illustrated in Fig. 3, the process of extracting prior information can be delineated into several stages: Gaussian filtering, edge detection, Hough line detection \cite{ref46}, and subsequent intersection calculation.

Initially, the module undertakes Gaussian filtering on the trajectory image, which is represented in grayscale. This process is pivotal for enhancing the image's smoothness and minimizing interference during the gradient calculation essential for edge detection. We let $\bm{I_{o}}$ represent the original image, $\bm{I_{p}}$ represent the denoised image. The pixel coordinates within $\bm{I_{o}}$ are denoted by $(x, y)$. Then the normalized Gaussian kernel $\bm{Kernel_{n}}$ required for Gaussian filtering has a corresponding value $\bm{Kernel_{n}}(x,y)$. The calculation for each pixel makes up a complete Gaussian kernel for the image. The denoised image is obtained by convolution with the original image. The calculation formula is:

\begin{equation}
    \bm{Kernel}(x, y) = \frac{1}{2\pi\sigma^2} e^{-\frac{x^2 + y^2}{2\sigma^2}},
    \label{eq:gaussian_kernel}
\end{equation}
\begin{equation}
    \bm{Kernel_n}(x, y) = \frac{\bm{Kernel}(x, y)}{\sum_{x=0}^{1279} \sum_{y=0}^{649}\bm{Kernel}(x, y)},
    \label{eq:gaussian_kerneln}
\end{equation}
\begin{equation}
    \bm{I_p}(i, j) = \sum_{u=0}^{1279} \sum_{v=0}^{649} \bm{I_0}(u, v) \bm{Kernel_n}(i-u, j-v),
    \label{eq:convolution}
\end{equation}
where $\sigma$ is the variance of Gaussian distribution. 

Next, the denoised image is subjected to edge detection using the Canny algorithm \cite{ref47}, which facilitates the extraction of pixel coordinates corresponding to the edges present in the image. To initiate this process, we compute the pixel gradient matrices of $\bm{I_{p}}$ by employing the Sobel operators. The Sobel operators consist of two matrices: $\bm{S_{x}}$ and $\bm{S_{y}}$. We use $\bm{S_{x}}$ to calculate the gradient matrix $\bm{g_{x}}$ in the $x$ direction. $\bm{S_{y}}$ is used to calculate the gradient matrix $\bm{g_{y}}$ in the $y$ direction. The calculation process is expressed as follows:

\begin{equation}
    \bm{g_x} = \bm{S_x} * \bm{I_p}; \quad \bm{g_y} = \bm{S_y} * \bm{I_p},
\end{equation}
\begin{equation}
 \bm{G}(x, y) = \sqrt{\bm{g_x}^2(x, y) + \bm{g_y}^2(x, y)},
 \label{eq:gradient_magnitude}
\end{equation}
\begin{equation}
    \bm{\theta}(x, y) = \arctan\left(\frac{\bm{g_y}(x, y)}{\bm{g_x}(x, y)}\right),
    \label{eq:angle}
\end{equation}
where $\bm{G}$ denotes the amplitude of the image gradient, while $\bm{\theta}$ signifies the direction of the image gradient. The symbol * is utilized to indicate the cross-correlation operation.

Once the above matrices are obtained, the Canny algorithm performs non-maximum suppression. This process retains only the local maxima of $\bm{G}$ based on $\bm{\theta}$, thereby narrowing the edges to a precise 1-pixel width. Then, we set two gradient magnitude thresholds, i.e., one high and one low. An edge is categorized as a strong edge if its gradient magnitude exceeds the high threshold, whereas an edge is identified as a weak edge if its gradient magnitude falls between the two thresholds. Thereafter, we retain only the strong edges and the weak edges that are connected to the strong edges to establish definitive edges.

Next, we determine the line equation of the edge using Hough line detection. In Hough line detection, the pixel $A(m_0, n_0)$ in the image space can represent a line $P_A$ in the parameter space. The line equation of $P_A$ is as follows:
\begin{equation}
    P_A: b = -m_0 a + n_0,
\end{equation}
where $a$ represents the abscissa of the parameter space (corresponding to the slope of a line passing through $A$ in image space) and $b$ represents the ordinate of the parameter space (corresponding to the intercept of this line in image space).
Consequently, the Hough transform allows us to convert edge pixel points into lines within the parameter space. The intersections of these lines facilitate the extraction of both the slope and intercept of the edges represented in image space. Subsequently, we can formulate the line equations for the edges identified in the preceding step.

Finally, we proceed to determine the corner points of the sideline based on the detected edges. Edge detection will yield a set of parallel edges. It is essential to merge these parallel and proximate edges into one edge. Then we selects two corner points as the prior information.

\subsection{Trajectory Classification Module}
\begin{figure*}[!t]
\centering
\subfloat[]{\includegraphics[width=3.5in]{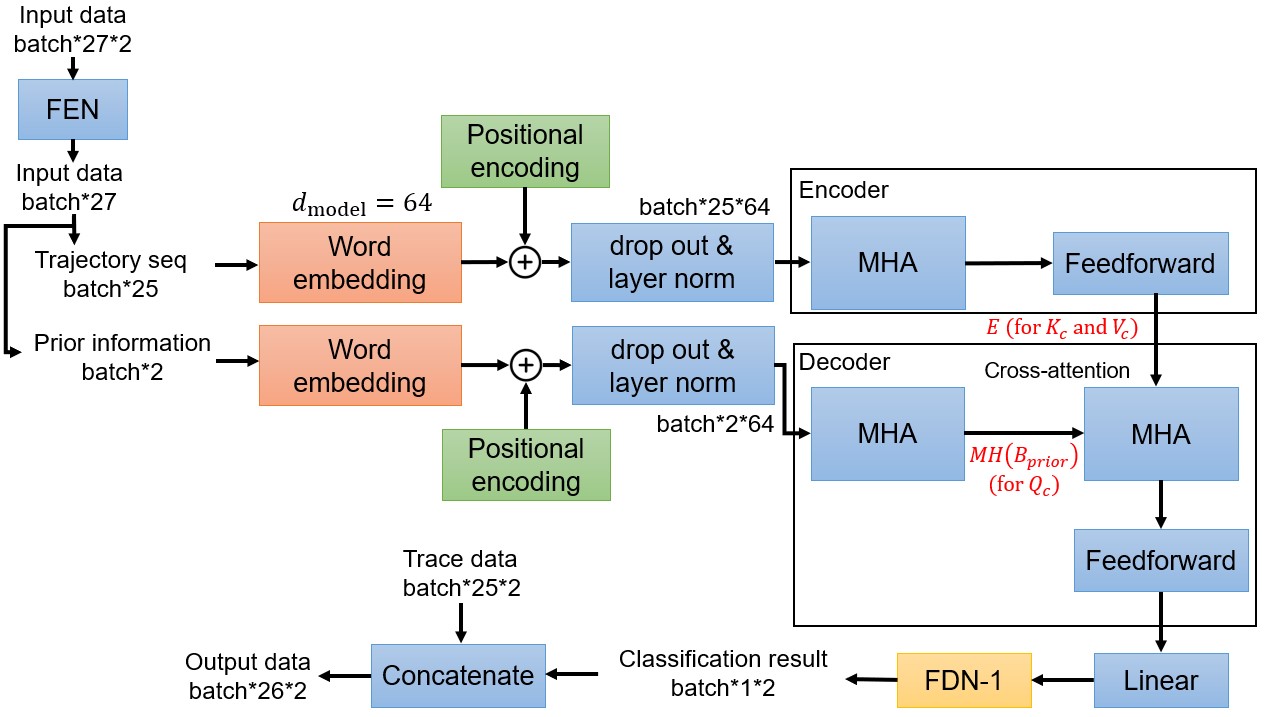}%
\label{fig4sub1}}
\hfil
\subfloat[]{\includegraphics[width=3.5in]{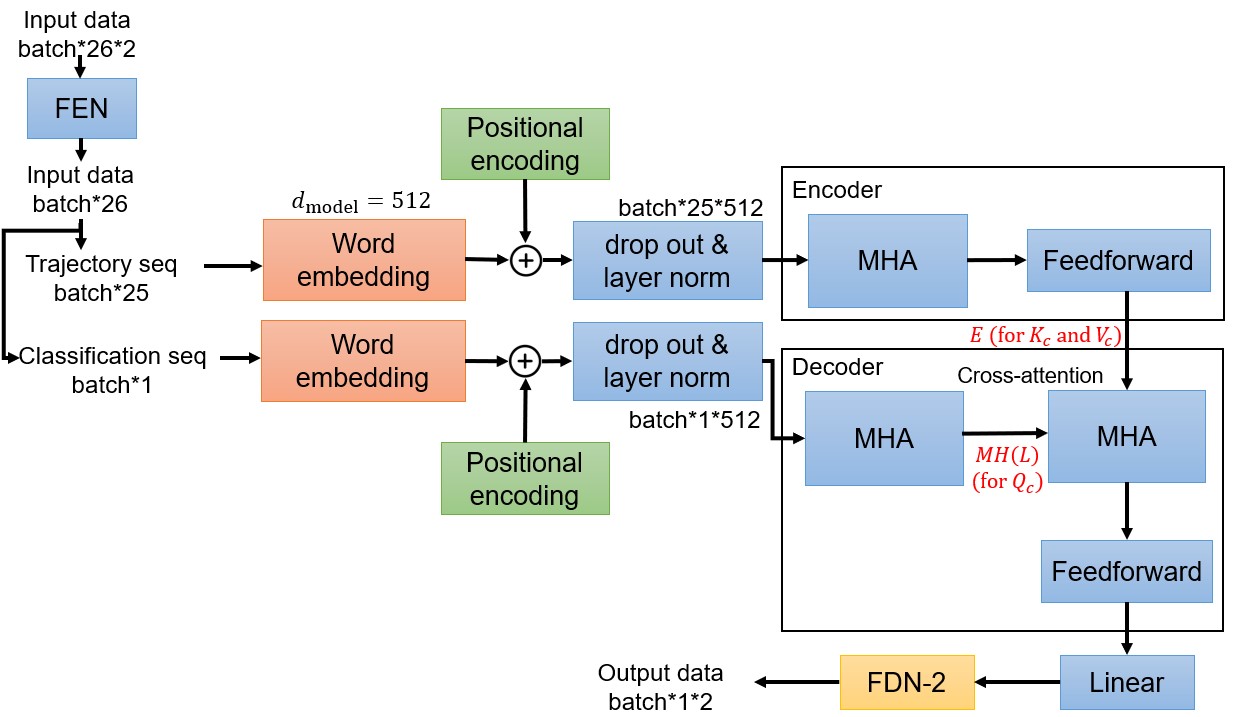}%
\label{fig4sub2}}
\caption{(a) Structure of the trajectory classification module. (b) Structure of the landing point prediction module.}
\label{fig4}
\end{figure*}
The trajectory classification module determines whether a flight path will land ``in'' or ``out'' of bounds by integrating sequential trajectory data with environmental priors. This fusion is achieved via a cross-attention mechanism. The module outputs discrete classification labels—an advantageous format that explicitly guides the subsequent prediction network, enabling it to learn and incorporate prior spatial contexts more effectively.  

Illustrated in Fig. 4(a), the Transformer-based classification module consists of an encoder that extracts dynamic trajectory features and a decoder that processes static prior information to predict the final classification. The module takes 25 trajectory points ($T_{ball}$) and 2 prior information points ($B_{prior}$) as input, generating a 1D vector ($Label$) as output. The initial input is formulated as:
\begin{equation}
    Input = Concat(T_{ball},B_{prior}),
\end{equation}
where $Concat()$ represents matrix concatenation operation. This direct concatenation facilitates subsequent sequence segmentation within the Feature Encoding Network (FEN), preventing premature fusion and preserving the temporal dynamics of the trajectory. Following dimensional transformation, the input is split into distinct trajectory and prior information sequences. After token embedding and positional encoding, these sequences are processed independently via Multi-Head Attention (MHA) before being fused through a cross-attention mechanism. Finally, the network computes the loss and performs backpropagation.

Since the trajectory classification module is based on a binary classification framework, the Binary Cross-Entropy (BCE) loss function has been selected for training this model, as presented below:
\begin{equation}
    BCE = -\frac{1}{N}\sum_{i=1}^{N}(q_i log(p_i) + (1-q_i)log(1-p_i)),
    \label{eq:BCE loss}
\end{equation}
where $q$ is the out-of-bounds label (0 or 1), $p$ is the predicted value from the trajectory classification module, and $N$ is the total number of trajectories.

The output $Label$ is concatenated with trajectory data to preserve the temporal dynamics of the trajectory, similar to the input structure. The concatenation process is given by:
\begin{equation}
    Output = Concat(T_{ball}, Label),
\end{equation}
\subsection{Landing Point Prediction Module}
Illustrated in Fig. 4(b), the landing point prediction module receives an input of 25 trajectory points alongside the classification $Label$, outputting the final predicted landing coordinates in 2D space. The input is initially flattened from 2D to 1D using a FEN comprising two linear layers activated by a Rectified Linear Unit (ReLU) function \cite{ref48}. The flattened data is then explicitly separated back into the trajectory sequence and the classification label. This division ensures that the subsequent positional encoding accurately annotates the trajectory order, thereby preserving its temporal dynamics. Following separation, token embedding projects both data streams into vectors of dimension $d_{model}$.

Positional encoding function is used to infuse the model with the necessary sequential context \cite{ref49}, which is expressed as
\begin{equation}
    PE(pos,2i) = sin(pos/10000^{2i/d_{model}}),
     \label{Positional code A}
\end{equation}
\begin{equation}
    PE(pos,2i+1) = cos(pos/10000^{2i/d_{model}}),
     \label{Positional code B}
\end{equation}
where $pos$ is the position of the element within trajectory sequence and $i$ is the dimension of positional code. The alternating expression of sine function and cosine function can mark the relative positions of different elements of the trajectory. We divide the data of $d_{model}$ dimensions into $d_{model}/2$ groups. Two data sequences in each group are respectively represented by $PE(pos,2i)$ and $PE(pos,2i+1)$.

Following positional encoding, the trajectory data $\bm{D}$ enters the encoder's multi-head attention mechanism \cite{ref49}. Here, the self-attention layer maps the input sequence into three distinct feature spaces—Queries, Keys, and Values—to compute the attention weights. These spaces are respectively defined by the query matrix $\bm{Q}$, key matrix $\bm{K}$, and value matrix $\bm{V}$, i.e.,
\begin{equation}
    \bm{Q} = \bm{D} \bm{W_Q}; \quad \bm{K} = \bm{D} \bm{W_K}; \quad \bm{V} = \bm{D} \bm{W_V},
\end{equation}
\begin{equation}
    Attention(\bm{Q},\bm{K},\bm{V}) = softmax(\frac{\bm{Q}\bm{K}^T}{\sqrt{d_k}})\bm{V},
    \label{Attention}
\end{equation}
where $\textbf{\textit{W}}$ denotes the corresponding weight matrix, $d_k$ denotes the length of the key vector. 

The results of self-attention are used to calculate the multi-head attention value, which is a combination of multiple self-attention value $head$, which is expressed as follows:
\begin{equation}
    head_i = Attention(\bm{QW^Q}_i,\bm{KW^K}_i,\bm{VW^V}_i),
    \label{head-Attention}
\end{equation}
\begin{equation}
    MH(\bm{D}) = Concat(head_1,head_2,...,head_h)\bm{W}^0,
    \label{Mulhead-Attention}
\end{equation}
where $MH(\bm{D})$ is the multi-head attention value of the trajectory, and $\bm{W}^0$ denotes the weight matrix. Consequently, the multi-head attention value is output from the feedforward module.

In the decoder, we also make attention calculations for the classification label data $\bm{L}$, resulting in the multi-head attention value $MH(\bm{L})$. Subsequently, we implement cross-attention mechanisms utilizing $MH(\bm{L})$ alongside the outputs from the encoder $\bm{E}$. Within the cross-attention framework, $MH(\bm{L})$ is used to calculate $\bm{Q}_c$, and $\bm{E}$ is used to calculate $\bm{K}_c$ and $\bm{V}_c$. Then we calculate multi-head attention value based on $\bm{Q}_c$, $\bm{K}_c$ and $\bm{V}_c$. $\bm{Q}_c$ provides the trajectory state (``in'' or ``out'') to influence the prediction area of the landing point, while $\bm{K}_c$ and $\bm{V}_c$ supply the temporal features of the trajectory for influencing the precise position of the prediction point. The resulting cross-attention output is then processed through a feedforward module, resulting in the generation of the decoder output. 

Finally, we achieve the 2D predicted landing point from Feature Decoding Network-2 (FDN-2). It is worth noting that both FDN-2 and FDN-1 consist of two linear layers; however, the key difference is that FDN-1 requires normalization while FDN-2 does not; and the activation functions for FDN-1 and FDN-2 are Sigmoid and ReLU, respectively.

We choose Mean Squared Error (MSE) as the loss function:

\begin{equation}
    MSE = \frac{1}{N}\sum_{i=1}^{N}(truth_i - prediction_i)^2,
    \label{eq:MSE loss}
\end{equation}
where $truth$ is the landing point coordinate, $prediction$ is the predicted value of landing point prediction module, and $N$ is the total number of trajectories.

\section{Experimental Results}
\subsection{Implementation Details}
\begin{table}[h]
\caption{Parameters for training\label{tab:table2}}
\centering
\begin{tabular}{|c||c||c|}
\hline
Parameter & Classification & Prediction\\
\hline
Epoch & 500 & 1000\\
\hline
Batch-size & 10 & 10\\
\hline
Learning rate & 0.0001 & 0.0001\\
\hline
Embedding number & 128 & 500\\
\hline
$d_{model}$ & 64 & 512\\
\hline
Dropout rate & 0.1 & 0.1\\
\hline
Encoder/Decoder layers & 1 & 1\\
\hline
Attention heads & 2 & 2\\
\hline
Feedforward neuron number & 256 & 2048\\
\hline
\end{tabular}
\end{table}

All experiments were implemented in Python using the PyTorch framework on a workstation equipped with an Nvidia GeForce RTX 3080 GPU. The dataset was partitioned into training and testing sets at a 4:1 ratio. We trained the proposed model using a batch size of 10 and the Adam optimizer, retaining the checkpoint that achieved the lowest validation loss for final testing. Specific training parameters for the different modules are detailed in Table II. In total, the finalized model contains 5.53M parameters, comprising 0.15M in the classification module and 5.38M in the prediction module.
\subsection{Evaluation Metrics}
For the classification module, we have selected $BCE$, $Accuracy$, $Precision$, and $Recall$ to conduct performance evaluation. For the prediction module, we have selected $MSE$, $RMSE$, and $Bias$ for evaluation purposes.

The $BCE$ loss function is used to evaluate the quality of the prediction results of a binary classification model. Its mathematical formulation is given in Equation (9).

$Accuracy$ represents the proportion of samples that have been accurately predicted:
\begin{equation}
    Accuracy = \frac{TP+TN}{TP+FP+TN+FN} \times 100\%,
    \label{eq:Accuracy}
\end{equation}
where $TP$ is the number of correctly predicted positive samples, $TN$ is the number of correctly predicted negative samples, $FP$ is the number of incorrectly predicted positive samples, and $FN$ is the number of incorrectly predicted negative samples.

$Precision$ denotes the proportion of accurately identified positive samples relative to the total number of samples classified as positive:
\begin{equation}
    Precision = \frac{TP}{TP+FP} \times 100\%.
    \label{eq:Precision}
\end{equation}

$Recall$ denotes the proportion of accurately identified positive samples relative to the total number of genuine positive samples:
\begin{equation}
    Recall = \frac{TP}{TP+FN} \times 100\%.
    \label{eq:Recall}
\end{equation}

$MSE$ represents the mean square error, which quantifies the discrepancy between the predicted landing point and the true landing point.

$RMSE$ represents the root mean square error between the predicted landing point and the true landing point:
\begin{equation}
    RMSE = \sqrt{\frac{1}{N}\sum_{i=1}^{N}(truth_i - prediction_i)^2}.
    \label{eq:RMSE}
\end{equation}

$Bias$ represents the average bias observed between the predicted landing point and the actual landing point. This metric is critical as it quantifies the systematic error in predictions, enabling a more comprehensive understanding of model accuracy and reliability in trajectory assessments:
\begin{equation}
    Bias = \frac{1}{N}\sum_{i=1}^{N}(truth_i - prediction_i).
    \label{eq:Bias}
\end{equation}

To providing a mapping from the pixel error to the physical error, we need to estimate the physical coordinate of the landing point $(x_{phy},y_{phy},z_{phy})$. Since all the predicted landing points are on the court surface (i.e., $z_{phy}=0$), the mapping process can be simplified to a 2D-2D conversion. Therefore, the pixel coordinates $(x_{img},y_{img})$ and the physical coordinates $(x_{phy},y_{phy})$ can be converted through the following homography matrix $\bm{H}$:
\begin{equation}
\begin{bmatrix}
x_{img} \\ y_{img} \\ 1
\end{bmatrix} 
= \bm{H}
\begin{bmatrix}
x_{phy} \\ y_{phy} \\ 1
\end{bmatrix}.
    \label{eq:homography matrix}
\end{equation}
It can be designed based on at least 4 points with known physical coordinates (e.g., the intersection points of the court sidelines). We use 10 points to calculate a more accurate $\bm{H}$. Then, the physical bias between the predicted coordinate $(x_i,y_i)$ and the actual one $(\hat{x}_i,\hat{y}_i)$ can be calculated as
\begin{equation}
    PhyBias = \frac{1}{N}\sum_{i=1}^{N}\sqrt{(x_i-\hat{x}_i)^2+(y_i-\hat{y}_i)^2}.
    \label{eq:Physical Bias}
\end{equation}

\subsection{Dataset Preprocessing}
To rigorously assess the efficacy of our proposed prediction method, we developed a comprehensive trajectory dataset, detailed in section III. Utilizing this dataset, we trained the prediction model as described in section IV. 

To train the trajectory classification module, the dataset preprocessing process involves the following 2 steps: (1) For each trajectory, we use the prior information extraction module to obtain two corner points as the prior information; (2) We compute the sideline based on prior points, then we determine the true classification label based on the positional relationship between the landing point and the sideline. The label is assigned a value of ``0'' when the classification outcome is identified as ``out''. if the classification result is ``in'', the label is set to ``1''. 

For the landing point prediction module, we incorporated these classification labels into the original trajectory dataset. This preprocessing approach enhances training efficiency by providing contextual information that aids in the learning process. 

\subsection{Test Set Experiment}
\begin{figure*}[!t]
\centering
\subfloat[]{\includegraphics[width=3.5in]{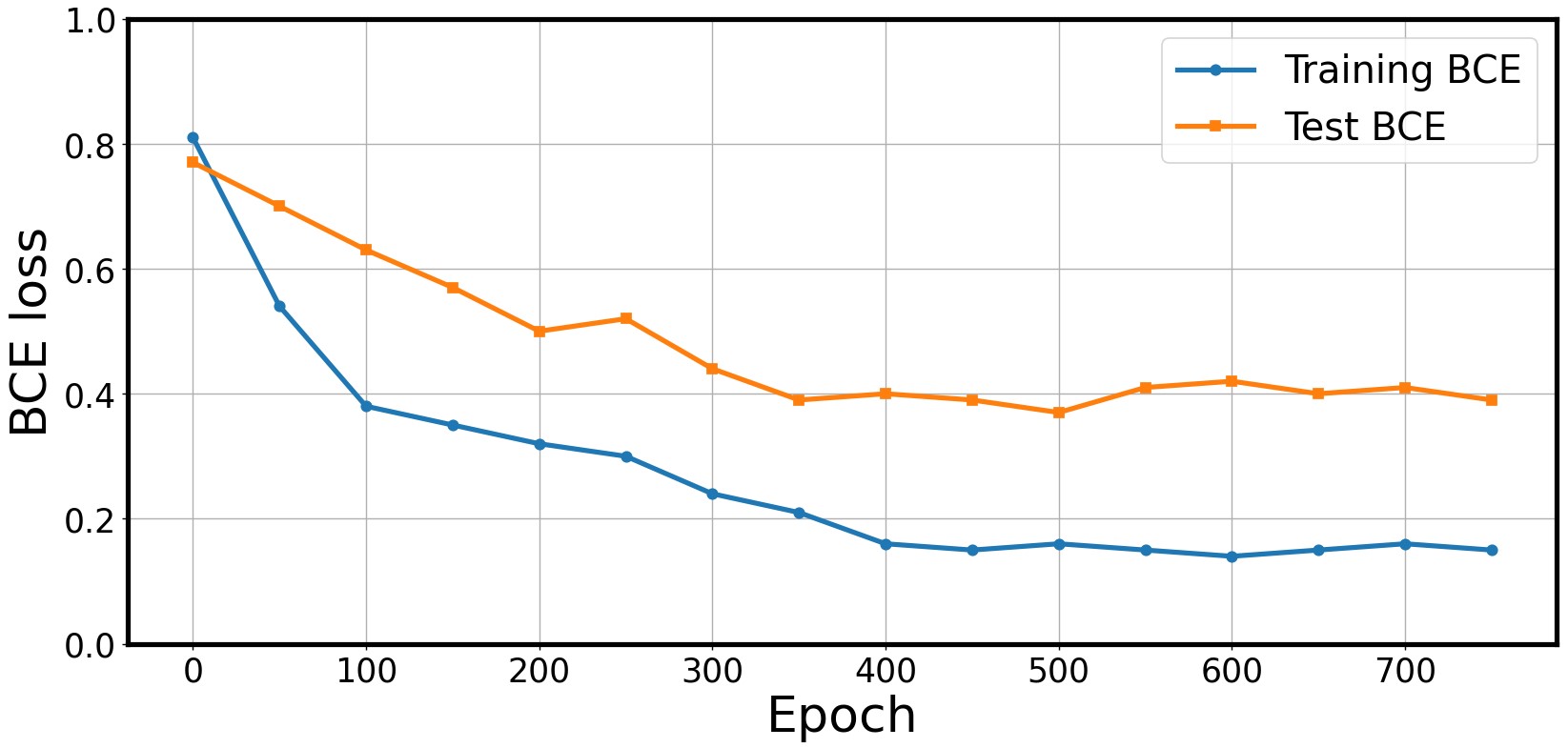}%
\label{fig5sub1}}
\hfil
\subfloat[]{\includegraphics[width=3.5in]{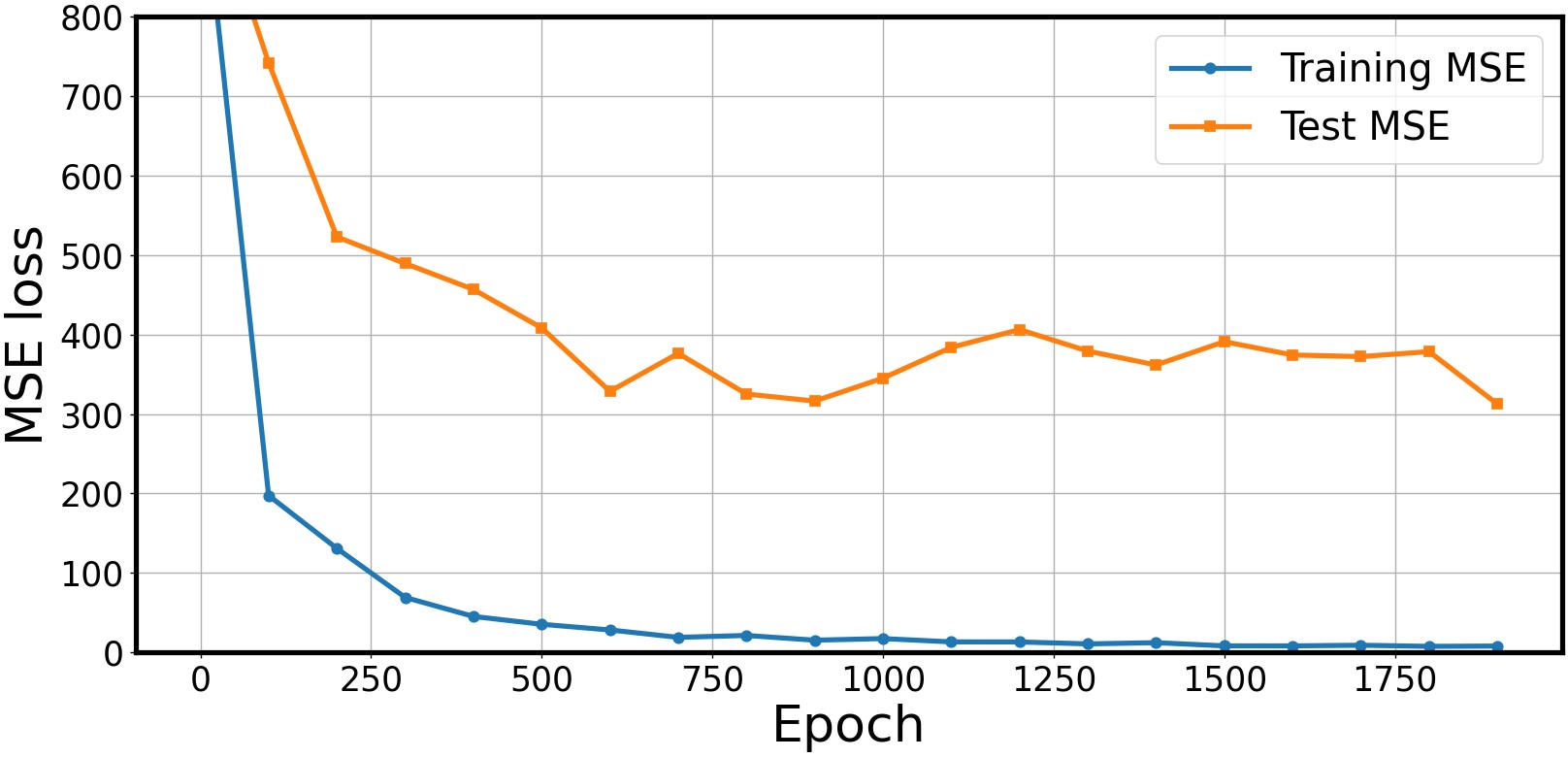}%
\label{fig5sub2}}
\caption{The loss during training and testing processes. (a) The BCE loss of classification module. (b) The MSE loss of prediction module.}
\label{fig5}
\end{figure*}

To rigorously assess the convergence of training and testing loss, the BCE and MSE associated with training and testing processes are illustrated in Fig. 5.

Figure 5(a) illustrates the variation of BCE loss as the number of epochs increases. The results demonstrate that classification module achieves effective convergence on test set. Figure 5(b) illustrates the variation of MSE loss as the number of epochs increases. The prediction module achieves similar effective convergence on the test set. Through comparative analysis, it is observed that the convergence speed of the prediction module is slower than that of the classification module. This is primarily due to the significantly large number of parameters in the prediction module.

\subsection{Ablation Experiment}

\begin{figure*}[!t]
\centering
\subfloat[]{\includegraphics[width=3.5in]{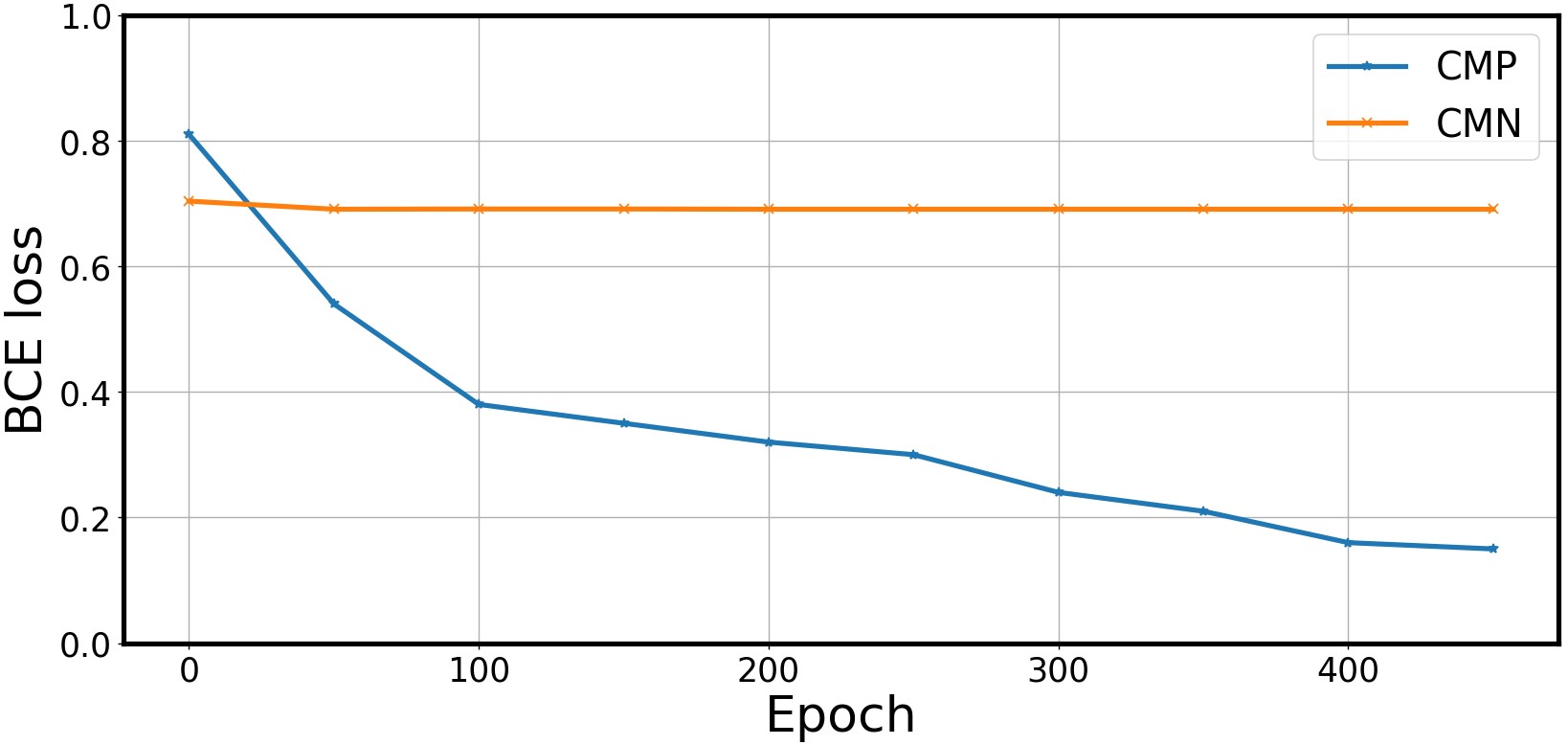}%
\label{fig6sub1}}
\hfil
\subfloat[]{\includegraphics[width=3.5in]{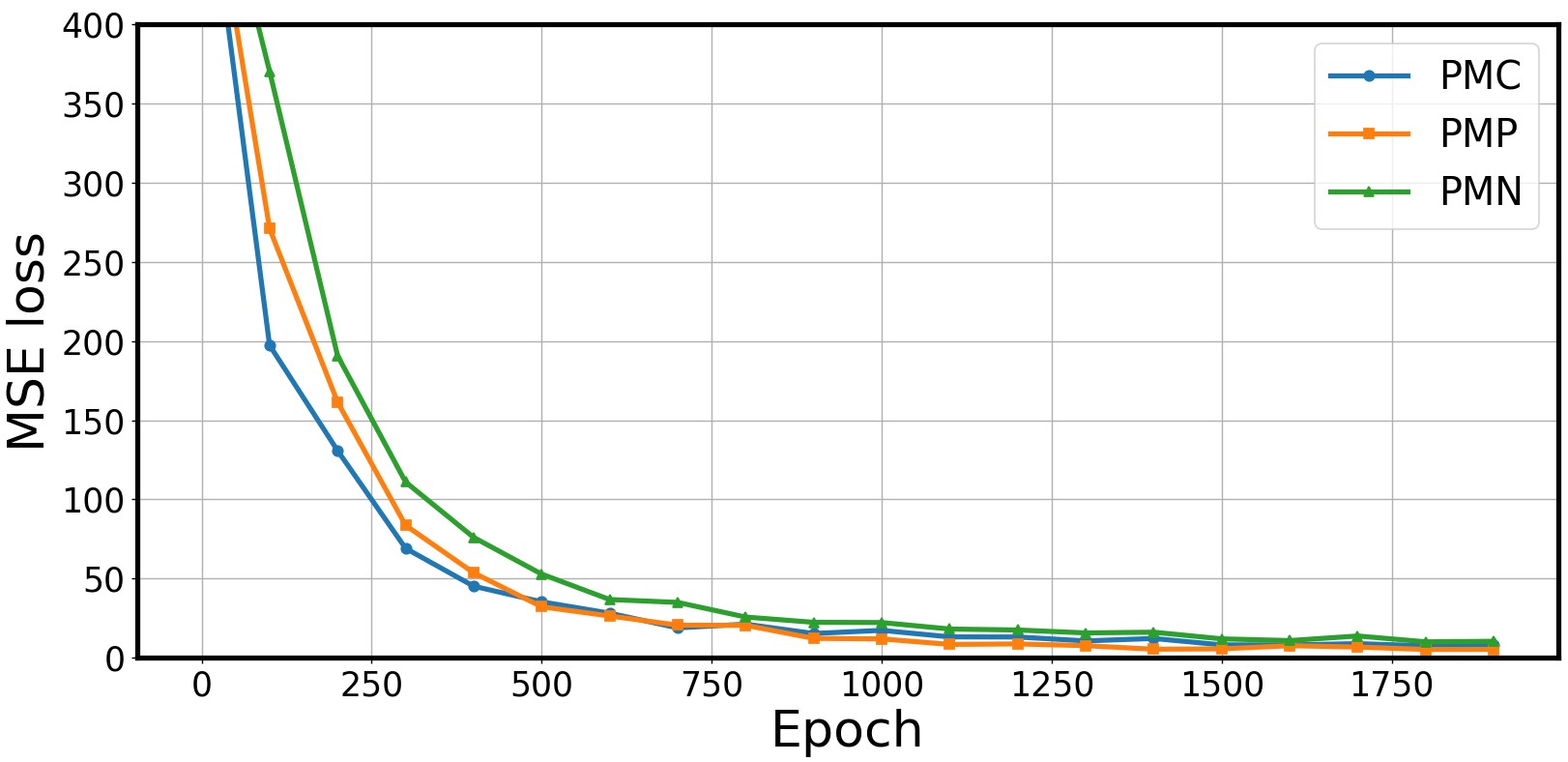}%
\label{fig6sub2}}
\caption{The loss of ablation models versus training epochs. (a) The BCE loss. (b) The MSE loss.}
\label{fig6}
\end{figure*}

\begin{table}[h]
\caption{Classification performance comparison of ablation models \label{tab:table3}}
\centering
\begin{tabular}{|c||c||c|}
\hline
 & CMN & CMP\\
\hline
Accuracy & 52.86\% & 85.71\%\\
\hline
Precision & 52.85\% & 81.40\%\\
\hline
Recall & 100\% & 94.59\%\\
\hline
\end{tabular}
\end{table}

\begin{table}[h]
\caption{Prediction performance comparison of ablation models \label{tab:table4}}
\centering
\begin{tabular}{|c||c||c||c|}
\hline
 & PMN & PMP & PMC\\
\hline
MSE & 1183.39 & 690.16 & \textbf{372.39}\\
\hline
RMSE & 34.40 & 26.27 & \textbf{19.30}\\
\hline
Bias (pixel) & 23.06 & 18.02 & \textbf{13.35}\\
\hline
PhyBias (cm) & 29.58 & 23.16 & \textbf{17.07}\\
\hline
\end{tabular}
\end{table}

To rigorously assess the effectiveness of our proposed prediction method, we conduct ablation experiments to examine the influence of various types of prior information on classification and prediction performance. The training losses associated with the different ablation models are illustrated in Fig. 6. The performance comparisons of ablation models are illustrated in Table III and IV.

For the classification ablation models, we denote the distinct types of the prior information related to null values and prior points as ``\textit{classification model with null value}'' (CMN) and ``\textit{classification model with prior information}'' (CMP), respectively. Figure 6(a) illustrates the variation of training BCE loss as the number of training epochs increases. The results demonstrate that CMP achieves effective convergence during training, in stark contrast to CMN, which fails to exhibit similar convergence behavior. Additionally, Table III provides a comprehensive overview of the classification performance across the ablation models when evaluated on the test set. The data presented therein reveals that only CMP exhibits effective classification capabilities. These findings highlight the importance of prior information points as essential features for trajectory classification. 

For the prediction ablation models, these models incorporate different forms of prior information for null values, prior points, and classification labels, referred to as ``\textit{prediction model with null value}'' (PMN), ``\textit{prediction model with prior information points}'' (PMP), and ``\textit{prediction model with classification labels}'' (PMC), respectively. Figure 6(b) illustrates the trend of the training MSE loss as training epochs progress. Notably, the PMC exhibits a more rapid convergence relative to the other ablation models under consideration. As detailed in Table IV, the predictive performance metrics of the various ablation models are presented. The PMC achieves the lowest losses across three evaluative criteria. Specifically, when compared to PMN, the PMC demonstrates reductions of 68.53\%, 43.90\%, and 42.11\% in the three criteria (MSE, RMSE, Bias). These results suggest that the integration of prior information significantly improves the model's predictive capabilities. Furthermore, when contrasted with the PMP model, the PMC persistently reveals lower values across all three criteria, underscoring that single classification label provide a more effective input than reliance solely on prior information points. 

\subsection{Comparative Experiments Across Different Learning Models}

\begin{figure*}[!t]
\centering
\subfloat[]{\includegraphics[width=3.5in]{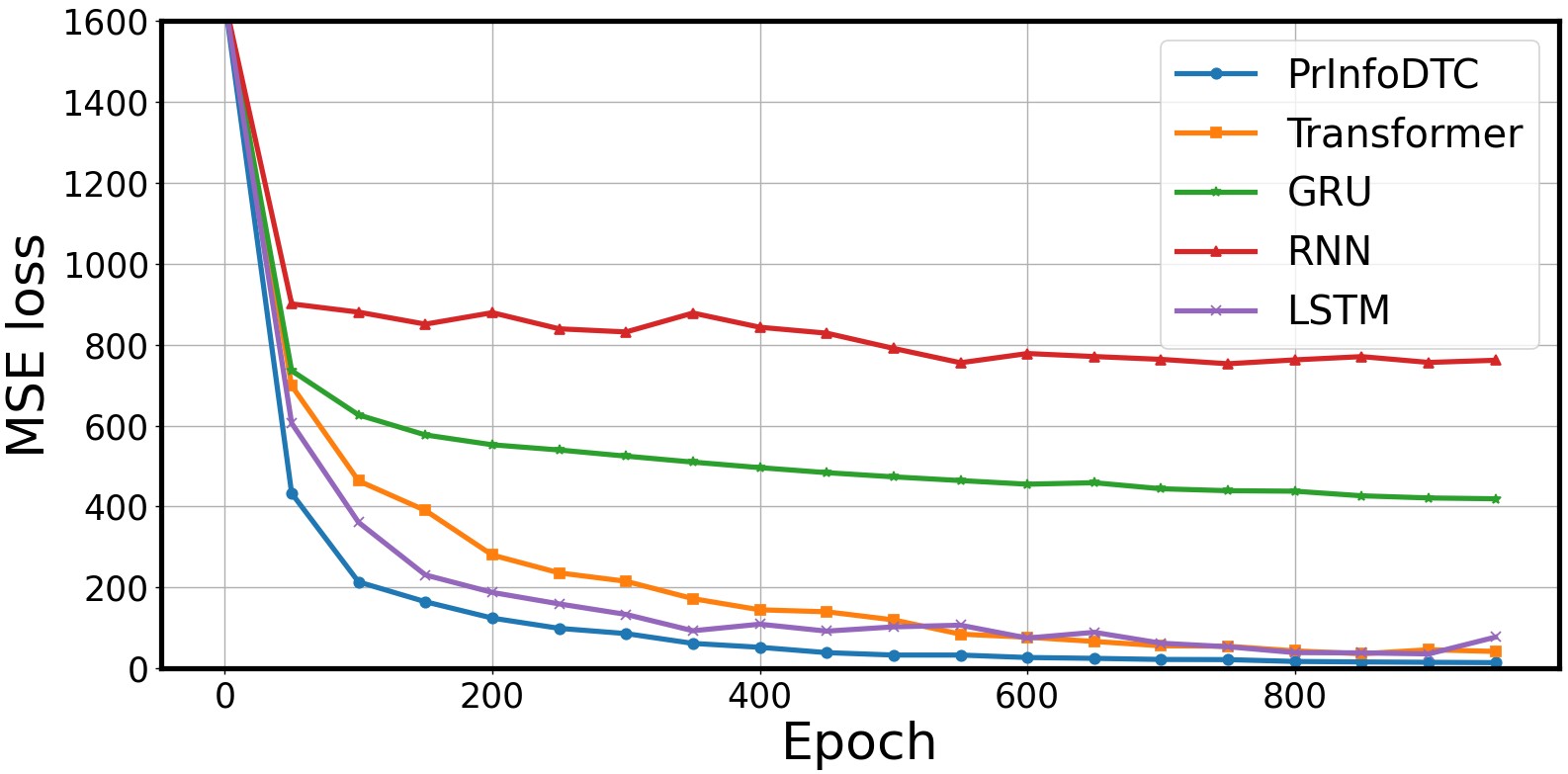}%
\label{fig7sub1}}
\hfil
\subfloat[]{\includegraphics[width=3.6in]{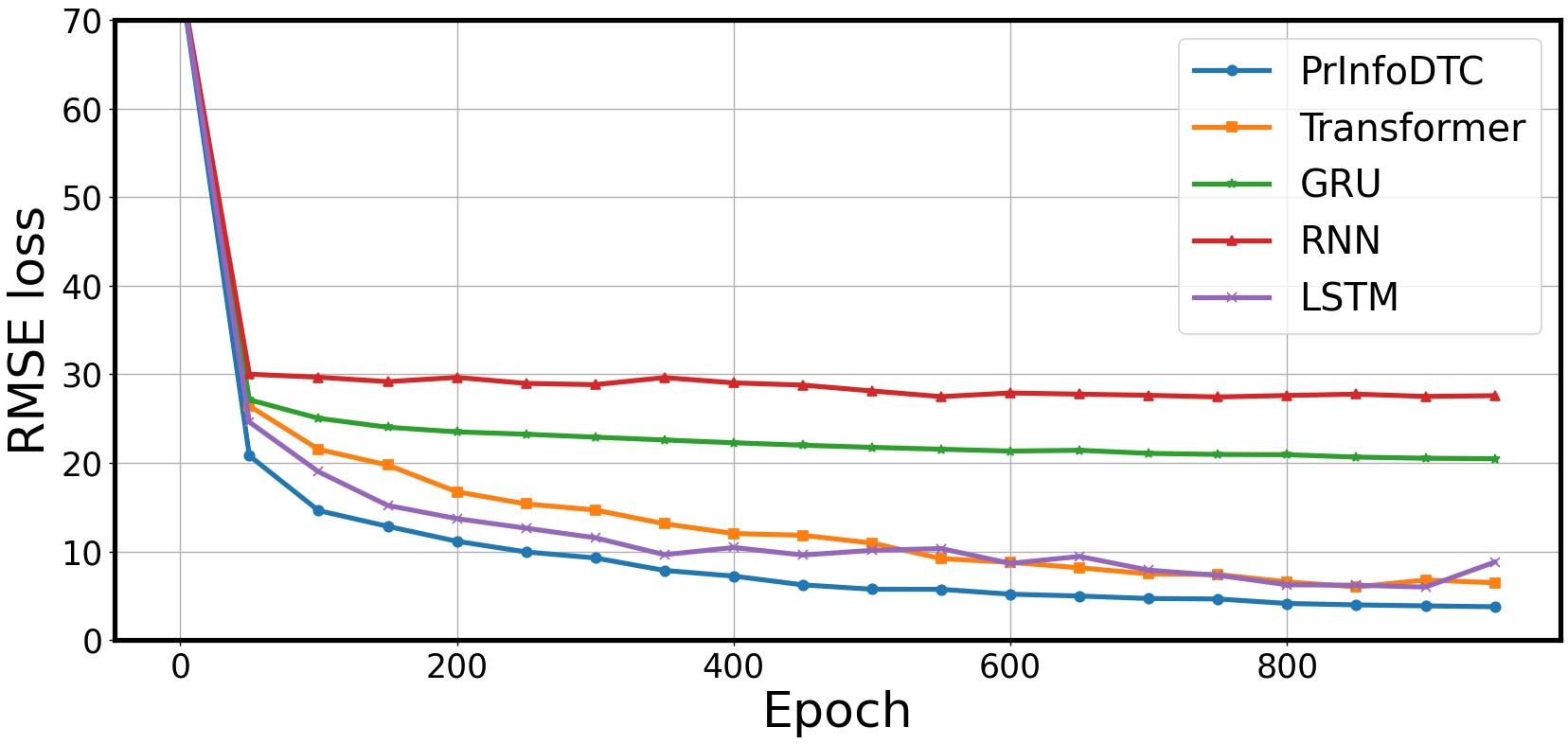}%
\label{fig7sub2}}
\\
\subfloat[]{\includegraphics[width=3.5in]{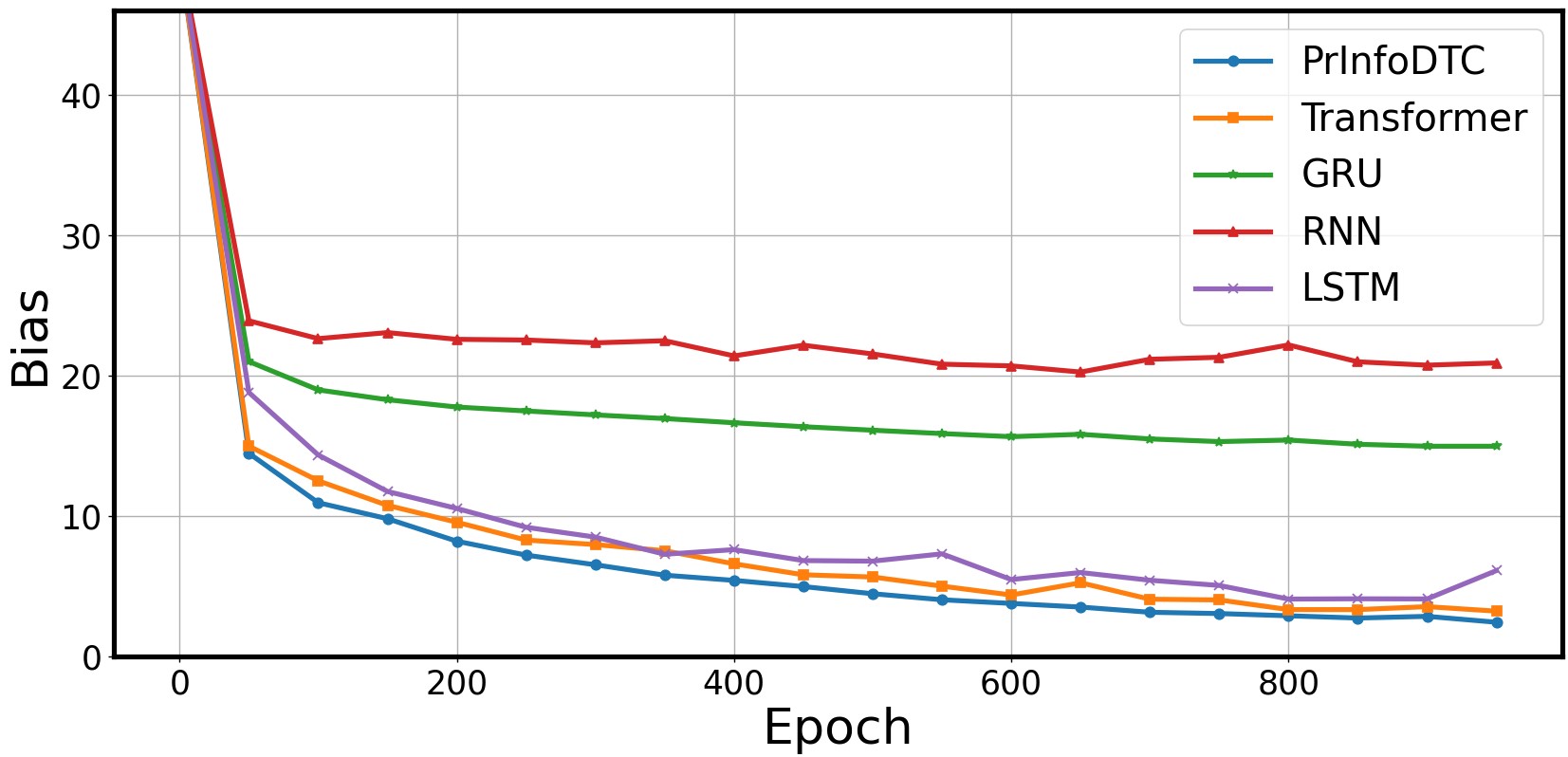}%
\label{fig7sub3}}
\caption{The training loss across different models versus training epochs: (a) MSE, (b) RMSE, (c) Bias.}
\label{fig7}
\end{figure*}

\begin{table}[h]
\caption{Prediction Performance Comparison ACROSS DIFFERENT MODELS \label{tab:table5}}
\centering
\scriptsize
\begin{tabular}{|c||c||c||c||c|}
\hline
 & MSE & RMSE & Bias (pixel) & PhyBias (cm)\\
\hline
RNN \cite{ref13} & 1064.99 & 32.63 & 26.71 & 34.16\\
\hline
GRU \cite{ref38} & 3417.77 & 58.46 & 49.24 & 63.98 \\
\hline
LSTM \cite{ref11} & 866.72 & 29.44 & 23.96 & 30.55 \\
\hline
Transformer \cite{ref35} & 1170.42 & 34.21 & 22.48 & 27.74\\
\hline
PIDTC & \textbf{372.39} & \textbf{19.30} & \textbf{13.35} & \textbf{17.07}\\
\hline
\end{tabular}
\end{table}

To illustrate the superior predictive capabilities of the proposed model, a comprehensive set of comparative experiments was carried out, benchmarked against established models used in previous research. For this comparison, four models were selected: RNN \cite{ref13}, GRU \cite{ref38}, LSTM \cite{ref11} and original Transformer \cite{ref35}. The results pertaining to three distinct loss metrics are illustrated in Fig. 7. Compared to the other predictive models, our proposed PIDTC model exhibits not only a faster convergence rate but also achieves a lower training loss at convergence.

Table V further shows the prediction performance across different models. Our proposed  model exhibits the lowest loss across all the three evaluative criteria, highlighting its effectiveness. While the basic Transformer achieves a lower bias compared to other standard models, its MSE loss remains higher than that of both RNN and LSTM models. These evaluation results imply that the structural attributes of the Transformer may enhance regression capabilities, though this improvement may come at the cost of precision in predicting certain specialized samples. The proposed PIDTC, featuring a dual-transformer-cascade structure—one dedicated to trajectory classification that incorporates prior information and the other focused on landing point prediction—effectively addresses the limitations associated with the standard Transformer.

\subsection{The Impact of Different Sizes of Training Set}

\begin{table}[h]
\caption{Prediction performance comparison for different training set sizes\label{tab:table6}}
\centering
\begin{tabular}{|c||c||c||c||c|}
\hline
 & 20\% $N_{t}$ & 40\% $N_{t}$ & 60\% $N_{t}$ & 80\% $N_{t}$\\
\hline
MSE & 499.41 & 547.52 & 542.15 & \textbf{372.39}\\
\hline
RMSE & 22.35 & 23.40 & 23.28 & \textbf{19.30}\\
\hline
Bias (pixel) & 15.65 & 17.10 & 16.22 & \textbf{13.35}\\
\hline
PhyBias (cm) & 19.98 & 21.57 & 20.65 & \textbf{17.07}\\
\hline
\end{tabular}
\end{table}

In addition, comparative experiments are conducted to evaluate the impact of varying the training set sizes, based on the same dataset that has $N_t=350$ samples. The proposed PIDTC model is trained using four distinct sizes of the training set, specifically 20\%$N_t$, 40\%$N_t$, 60\%$N_t$, and 80\%$N_t$. Subsequently, we assess the performance on the same test set and present the results in Table VI. The result reveals that the loss of the proposed model generally decreases as the size of the training set increases. 

\section{Conclusion}
In this paper, we presented a novel learning architecture for flying trajectory prediction based on prior environmental information and a cascaded transformer structure. To develop the flying trajectory dataset, we constructed an effective trajectory data acquisition platform that comprises a single 2D industrial camera and a ball launch machine. Our proposed model comprised three distinct sub-modules: a prior information extraction module for the identification of critical prior points, a trajectory classification module for ``in'' versus ``out'' classifications with respect to the court boundary, and a landing point prediction module dedicated to predicting the landing point coordinates. Subsequently, we calculated the sideline using Hough line detection and select two corner points as the relevant prior information. The trajectory classification module was then employed to generate classification labels. In the final step, trajectories and their corresponding classification labels were input into the landing point prediction module. Our ablation experiments validated the efficacy of the proposed approach. Comparative experiments reveal that our approach outperforms existing trajectory prediction frameworks, achieving a remarkable reduction in MSE/RMSE loss and Bias. Inspired by this work, future research endeavors will focus on integrating additional prior environmental information and adopting a physical-informed learning methodology.





\vfill

\end{document}